\documentclass{article}
\usepackage{aaai25}  
\usepackage{times}
\usepackage{helvet} 
\usepackage{courier} 
\usepackage[hyphens]{url} 
\usepackage{graphicx} 
\usepackage{natbib} 
\usepackage{caption} 
\usepackage{algorithm}
\usepackage{algorithmic}
\usepackage{booktabs}
\usepackage{multirow}
\usepackage{amsfonts}
\usepackage{amsmath}
\usepackage{tabularx}
\usepackage{subcaption}
\usepackage{float}
\usepackage{placeins}
\usepackage[dvipsnames]{xcolor} 
\usepackage[pagebackref=true,breaklinks=true,colorlinks,bookmarks=false,citecolor=RoyalBlue]{hyperref}

\usepackage{newfloat}
\usepackage{listings}
\DeclareCaptionStyle{ruled}{labelfont=normalfont,labelsep=colon,strut=off} 
\floatstyle{ruled}
\newfloat{listing}{tb}{lst}{}
\floatname{listing}{Listing}

\makeatletter
\def\@fnsymbol#1{\ensuremath{\ifcase#1\or \dagger\or *\or \ddagger\or
   \mathsection\or \mathparagraph\or \|\or **\or \dagger\dagger
   \or \ddagger\ddagger \else\@ctrerr\fi}}
   
\title{ConVis: Contrastive Decoding with Hallucination Visualization for \\ Mitigating Hallucinations in Multimodal Large Language Models
}
\author{
    Yeji Park\equalcontrib, Deokyeong Lee\equalcontrib, Junsuk Choe\thanks{Corresponding authors.}, Buru Chang\footnotemark[2]
}
\affiliations{
    \large{Sogang University}\\
    \texttt{\{yjparkm, plmft, jschoe, buru\}@sogang.ac.kr}
}

\usepackage{bibentry}
\usepackage{kotex}
\usepackage{caption}
\usepackage{xspace}

\begin{document}

\maketitle

\begin{abstract}
Hallucinations in Multimodal Large Language Models (MLLMs) where generated responses fail to accurately reflect the given image pose a significant challenge to their reliability.
To address this, we introduce ConVis, a novel training-free contrastive decoding method. 
ConVis leverages a text-to-image (T2I) generation model to semantically reconstruct the given image from hallucinated captions. 
By comparing the contrasting probability distributions produced by the original and reconstructed images, ConVis enables MLLMs to capture visual contrastive signals that penalize hallucination generation.
Notably, this method operates purely within the decoding process, eliminating the need for additional data or model updates.
Our extensive experiments on five popular benchmarks demonstrate that ConVis effectively reduces hallucinations across various MLLMs, highlighting its potential to enhance model reliability.
Source code is available at \url{https://github.com/yejipark-m/ConVis}
\end{abstract}
\section{Introduction}\label{sec:1_introduction}
Multimodal Large Language Models (MLLMs)~\cite{dai2023instructblip,liu2024visual} are advanced language models capable of understanding both images and text, such as image captioning and visual question answering (VQA).
While MLLMs have achieved significant success that utilize both visual and textual information, the issue of \textit{hallucination}, where the models generate responses that do not align with the given image, has greatly undermined their reliability~\cite{liu2023mitigating,sun2024aligning}.
This problem poses a significant obstacle to adopting MLLMs in critical fields where reliability is crucial. 
For instance, in medical applications, it could lead to incorrect diagnoses~\cite{liu2023medical}, while in MLLM-based autonomous systems, it might result in erroneous interpretations~\cite{shao2024lmdrive}.

\begin{figure}[t]
  \centering
  \includegraphics[width=\columnwidth]{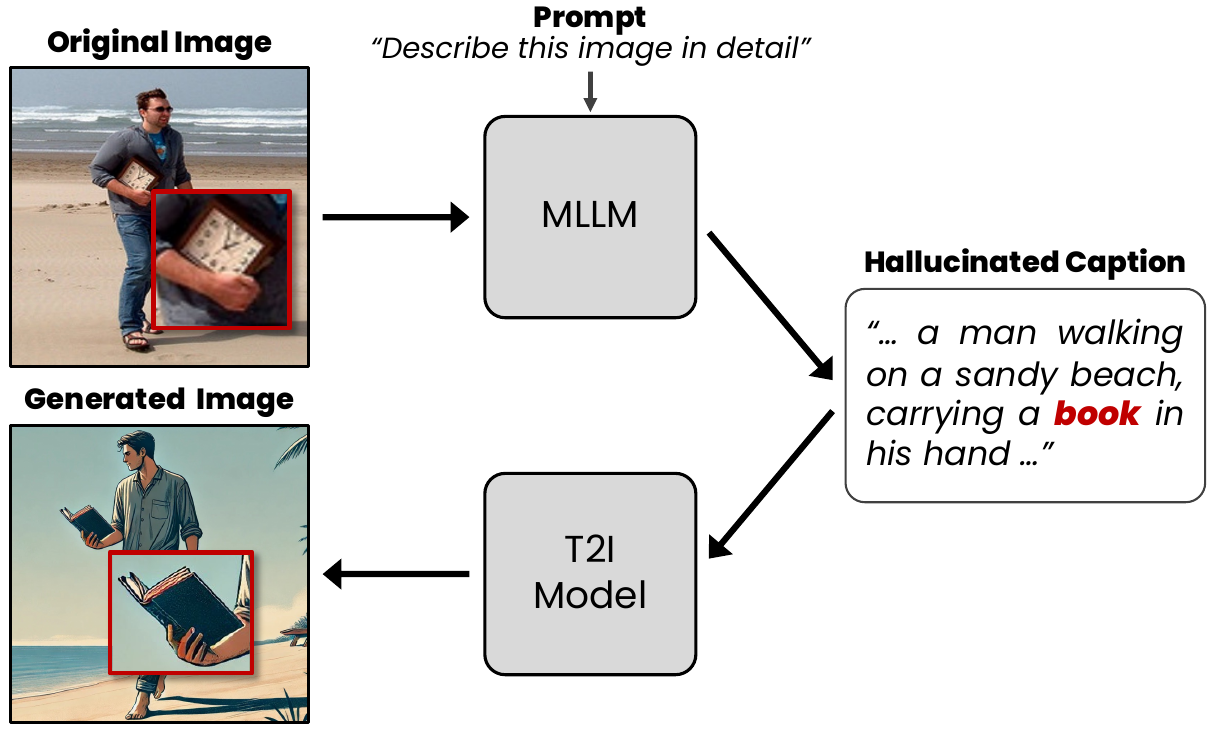}
  \caption{The text-to-image model visualizes hallucinations (\textit{e.g.}, `\textit{book}') in the semantically reconstructed images based on the hallucinated caption, exhibiting differences (\textit{e.g.}, missing `\textit{clock}') from the original image.}
  \label{fig:1_motivation}
\end{figure}

Recent research has been actively conducted to address this.
WoodPecker~\cite{yin2023woodpecker} and LURE~\cite{zhou2024analyzing} reduce hallucinations by post-processing the generated responses. 
Datasets such as LRV-Instruction~\cite{liu2023mitigating} and RLHF-V~\cite{yu2024rlhf} have been proposed to mitigate hallucinations through instruction tuning of MLLMs.
However, these studies often rely on external APIs like GPT-3.5, require costly human feedback collection, and necessitate additional training of MLLMs.

In contrast, this paper focuses on decoding strategies that reduce hallucinations by intervening solely in the decoding process, without the need for additional data or model training. 
The following studies fall into this category:
OPERA~\cite{huang2024opera} imposes penalties on token generation that does not reference visual tokens.
VCD~\cite{leng2024mitigating} creates contrasting distributions using distorted images to reduce the model's reliance on statistical biases and priors that lead to hallucinations.
HALC~\cite{chen2024halc} corrects hallucinations by leveraging cues provided by visual information from various fields of view.

\begin{figure}[t]
  \centering
  \includegraphics[width=\columnwidth]{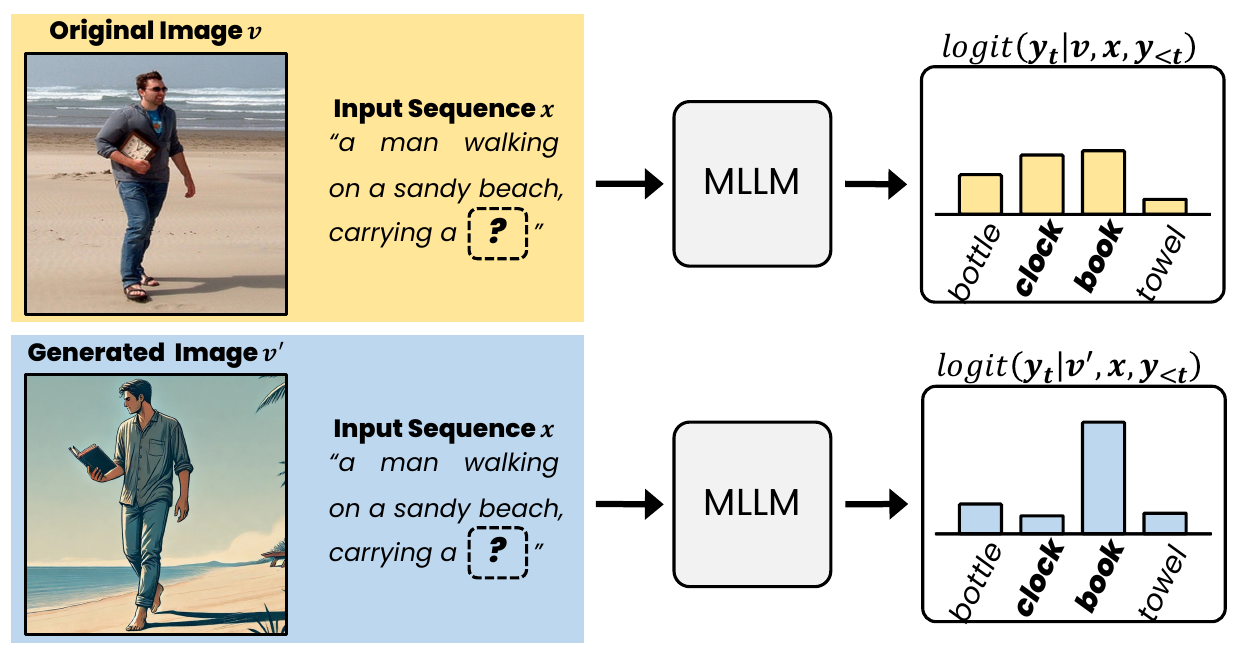}
  \caption{The original and reconstructed image generate the contrastive logit distribution for the hallucinated tokens (\textit{e.g.}, `\textit{book}'). The reconstructed image tends to amplify the logits of tokens corresponding to the visualized hallucination.}
  \label{fig:2_contrastive_distribution}
\end{figure}

In this study, we propose a contrastive decoding method called \textit{\textbf{ConVis}} (\underline{Con}trastive Decoding with Hallucination \underline{Vis}ualization), which can be applied to any existing MLLM without additional training.
Inspired by the previous work~\cite{kim2024exploiting}, ConVis leverages text-to-image (T2I) generation models, specifically Hyper-SDXL~\cite{ren2024hyper}, to capture visual contrast signals. The process begins with the MLLM generating a caption for the input image, after which the T2I model reconstructs an image based on this caption.
As shown in Figure~\ref{fig:1_motivation}, if the generated caption contains hallucinations (e.g., a \textit{book}'), there will be visual discrepancies between the original and reconstructed images (e.g., a missing \textit{clock}').
ConVis then uses the original and reconstructed images to compare the probability distributions (Figure~\ref{fig:2_contrastive_distribution}), capturing visual contrast signals that highlight hallucinations.
Based on these signals, ConVis penalizes the generation of hallucinations during the decoding process, reducing the hallucinations.

To validate the effectiveness of ConVis, we conducted experiments across five benchmarks: CHAIR~\cite{rohrbach2018object}, HallusionBench~\cite{guan2024hallusionbench}, POPE~\cite{li2023evaluating}, MME~\cite{fu2023mme} and LLaVA-Bench~\cite{liu2024visual}.
The results consistently demonstrated that our decoding method reduces hallucinations while maintaining overall response generation performance across various MLLMs, including LLaVA-1.5~\cite{liu2024improved}, MiniGPT-4~\cite{zhu2024minigpt}, and mPLUG-Owl2~\cite{ye2024mplug}.

Our contributions can be summarized as follows: 
(1) Propose ConVis, a novel contrastive decoding method that visualizes hallucinations using a T2I model. To the best of our knowledge, this is the first time a T2I model has been employed to mitigate hallucinations through a decoding strategy.
(2) Conduct extensive experiments to validate the effectiveness of ConVis in reducing hallucinations.
(3) Provide insights into how T2I models can serve as a valuable source of visual contrastive signals in decoding methods aimed at mitigating hallucinations.
\section{Related Work}\label{sec:2_related_work}
\subsection{Multimodal Large Language Models}\label{subsec:2_1_multimodal_large_language_models}
The emergence of LLMs has revolutionized the paradigm of Natural Language Processing (NLP). 
The significant success of LLMs in the NLP field has led to research on leveraging LLMs in the visual domain. 
Consequently, MLLMs that can simultaneously handle visual and textual data have recently been proposed.
Specifically, to process visual information, LLaVA~\cite{liu2024visual} uses a CLIP vision encoder~\cite{radford2021learning} and a linear layer to project images into the LLM's input embedding space. 
MiniGPT-4~\cite{zhu2024minigpt} employs a Q-Former~\cite{li2023blip} and a linear layer to project images into the LLM's input embedding space. 
Additionally, mPLUG-Owl2~\cite{lai2024lisa} introduces a modality-adaptive module that preserves modality-specific features, allowing the model to excel in both multimodal and NLP tasks.

However, despite these efforts, misalignment between modalities can still occur for various reasons, leading to generated responses that do not correspond to the visual information. 
This phenomenon, known as hallucination, undermines the reliability of MLLMs and poses a significant challenge to their application in real-world scenarios.

\subsection{Hallucination Mitigation}\label{subsec:2_2_hallucination_mitigation}
To address the hallucination problem in MLLMs, several studies have been proposed recently. 
Lure~\cite{zhou2024analyzing} and Woodpecker~\cite{yin2023woodpecker} employ post-processing methods to revise generated responses, either by training a revisor or using GPT-3.5-turbo~\cite{brown2020language}. 
Fine-tuning approaches~\cite{liu2023mitigating,yu2024rlhf} mitigate hallucinations through instruction tuning with additional data, but they require significant data collection and training resources. 
Given the large number of parameters in MLLMs, this is computationally inefficient.

Therefore, methods for improving the decoding process have recently received great attention due to the advantage that they do not require additional training. 
Specifically, OPERA~\cite{huang2024opera} explores aggregation patterns that cause hallucinations. 
OPERA utilizes this insight to suppress the generation of tokens that exhibit these patterns. 
VCD~\cite{leng2024mitigating} leverages the characteristic that the model tends to prioritize prior knowledge over visual information when responding to distorted images. As a result, the responses to the distorted image and the original image show significant differences in hallucinated tokens, and VCD contrasts these to mitigate the hallucinations. 
HALC~\cite{chen2024halc} observes that when images with varying fields of view are input into the MLLM, the probability changes for ground truth tokens are much greater than for hallucinated tokens. 
This observation helps identify visual context candidates that clearly depict objects, and by contrasting these candidates, HALC reduces hallucinations.

Unlike existing techniques, we propose a new decoding method that utilizes a T2I model. 
Specifically, our approach visualizes hallucinations in the initially generated caption using a T2I model, then contrasts the responses generated from the reconstructed image with those from the original image. 
Through this process, we contrast distributions of the hallucinated tokens and effectively mitigate hallucinations.

\section{Methodology}\label{sec:3_methodology}
\begin{figure*}[t]
  \centering  \includegraphics[width=\textwidth]{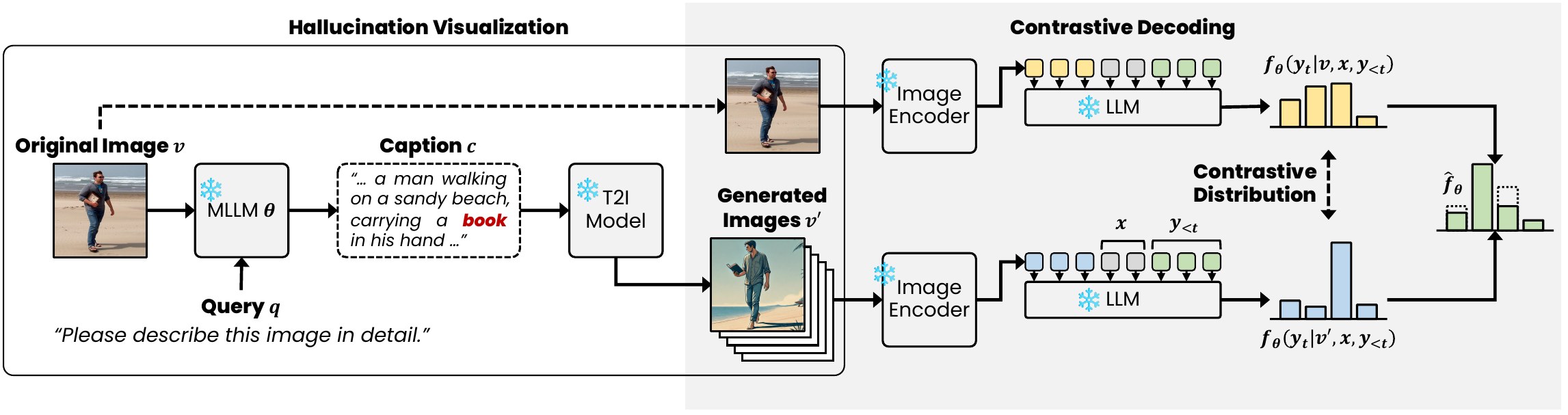}
  \caption{The original and generated image produce the contrastive distribution for the hallucinated tokens (\textit{e.g.}, `\textit{book}'). The generated image tends to amplify the logits of tokens corresponding to the visualized hallucination.}
  \label{fig:3_contrastive_distribution2}
\end{figure*}
\subsection{Preliminaries}\label{subsec:3_1_preliminaries}
\textbf{Response Generation.}
The MLLM generates a response \( y \) corresponding to a given input image \( v \) and instruction text \( x \).
The input image is projected into visual tokens through an image encoder, and these tokens, along with the tokens corresponding to the instruction text, are fed into the LLM. 
The response is generated through autoregressive decoding according to the following equation:
\begin{equation}
y_t \sim p_\theta(\cdot|v,x,y_{<t}) \propto \exp\big(f_\theta(\cdot|v,x,y_{<t})\big),
\label{eq:1_response_generation}
\end{equation}%
where \( \theta \) denotes the parameters of the MLLM, \( y_t \) represents the \( t \)-th token of response, and \( y_{<t} \) is the sequence of tokens generated up to time \( t \). \( f_\theta \) denotes the logit distribution generated by the MLLM. 
Hallucination refers to the phenomenon where the output \( y \) generated by the MLLM does not correspond to the input image \( v \). 
This study focuses on mitigating hallucinations while maintaining the overall performance of the MLLM as a language model.

\noindent
\textbf{Text-to-Image Generation.}
The core component of ConVis is the T2I model that generates images based on a given query. 
The goal of the T2I model is to create an image that accurately depicts the query. 
Among the recently proposed T2I models, we utilize Hyper-SDXL~\cite{ren2024hyper}, an enhanced version of Stable Diffusion~\cite{ho2020denoising}, which has demonstrated excellent T2I performance. 
The diffusion-based Hyper-SDXL model begins with a pure noise and progressively reconstructs it through an iterative reverse diffusion process which ultimately results in the generated image \( v'_0 \).

\subsection{Hallucination Visualization}\label{subsec:3_2_hallucination_visualization}
We hypothesize that the T2I model can help mitigate hallucinations by providing visual contrast signals during the decoding process. 
If the T2I model receives a caption generated by the MLLM that contains hallucinations, it will faithfully visualize those hallucinations in the generated image. 
We refer to this process as \textit{hallucination visualization}.

To implement this, ConVis first generates an initial caption \( c \) for the original image \( v \) using a simple instruction text that directs the MLLM to describe the image. This process is illustrated in Figure~\ref{fig:2_contrastive_distribution}. 
The T2I model then takes the caption \( c \) as a query and generates an image \( v' \) based on it. 
If the caption contains hallucinations, these will be faithfully visualized in the generated image \( v' \). 
Conversely, if the initial caption is accurate and free of hallucinations, the generated image will be semantically similar to the original image.

\noindent
\textbf{Diversity of Generated Images.}
Given that the current T2I model may not generate images that fully align with the captions, we address this limitation by increasing the diversity of the generated images using the following approaches:
(1) We first generate a diverse set of $n$ captions using Nucleus Decoding~\cite{holtzman2020curious} instead of Greedy Decoding. 
(2) Then, the T2I model uses these $n$ captions to generate $n$ corresponding images. 
This approach increases coverage of the various potential hallucinations that the MLLM might generate by diversifying the captions. 
Additionally, by using multiple images instead of a single one, we enhance the robustness of our method against the T2I model’s potential misalignment between the caption and the generated image due to its imperfect performance.

We have found these approaches to be effective, with detailed results available in the experiment section.

\subsection{Contrastive Decoding}\label{subsec:3_3_contrastive_decoding}
Hallucinations in captions cause visual differences between the original image $v$ and the generated image $v'$. 
We mitigate these hallucinations by capturing the visual contrast signals from these differences. 
To achieve this, during the decoding process, we utilize both the original image $v$ and the $n$ generated images to produce the logit distribution for each image. 
The final contrastive logit distribution $\hat{f_\theta}$ is derived by averaging the contrastive logit distributions between the original image and each generated image as follows:

\begin{equation}
\begin{aligned}
    \hat{f_\theta} &= \frac{1}{n}\sum_{i=1}^n{\Big((1+\alpha)f_\theta(\cdot|v,x,y_{<t}) \Big.} 
    -\alpha f_\theta(\cdot|v'_i,x,y_{<t})\Big),
\end{aligned}
\label{eq:3_contrastive_decoding}
\end{equation}%
where $\alpha$ is a hyperparameter that controls the strength of the difference between the logit distributions from the original and generated images.
The contrastive logit distribution $\hat{f_\theta}$ is used to generate the response $y$.
For tokens associated with hallucinations, the contrastive logit distribution is significantly amplified compared to other tokens, allowing us to penalize these tokens and reduce the hallucinations. 

\begin{table*}[ht!ht]
\centering
\begin{tabular}{@{}c|cc|cc|cc@{}}
\toprule
\multirow{2}{*}{Method} & \multicolumn{2}{c|}{LLaVA-1.5}             & \multicolumn{2}{c|}{mPLUG-Owl2}            & \multicolumn{2}{c}{MiniGPT-4}               \\ \cmidrule(l){2-7} 
                 & CHAIR$_S$ ↓        & CHAIR$_I$ ↓ & CHAIR$_S$ ↓        & CHAIR$_I$ ↓ & CHAIR$_S$ ↓        & CHAIR$_I$ ↓        \\ \midrule
Greedy Search          & 22.4 ± 1.11       & 7.4 ± 0.27 & 22.2 ± 1.10       & 7.3 ± 0.24 & 34.0 ± 1.11       & 13.8 ± 0.85       \\
Nucleus Sampling & 26.0 ± 1.93       & 9.5 ± 0.76 & 25.2 ± 1.59       & 9.3 ± 0.34 & 30.1 ± 1.45       & 14.2 ± 0.90       \\
Beam Search              & 19.5 ± 1.42          & \textbf{6.4 ± 0.09}    & 18.3 ± 0.42          & \textbf{6.0 ± 0.34} & 31.1 ± 1.03          & 12.4 ± 0.59          \\ \midrule
VCD              & 23.7 ± 1.90       & 8.2 ± 0.80 & 25.7 ± 1.30       & 9.0 ± 0.28 & 31.6 ± 1.83       & 13.8 ± 0.83       \\
OPERA            & \underline{18.5 ± 0.90} & 6.6 ± 0.23 & \underline{18.2 ± 0.40} & 6.2 ± 0.18 & 30.6 ± 1.06       & 12.5 ± 0.91       \\
HALC             & 23.7 ± 2.66       & 9.1 ± 0.41 & 24.3 ± 1.22       & 9.4 ± 0.19 & \underline{24.2 ± 1.91} & \underline{10.8 ± 0.53} \\ \midrule
Ours                    & \textbf{18.4 ± 0.53} & \underline{6.4 ± 0.37} & \textbf{17.6 ± 3.54} & \underline{6.0 ± 0.89}    & \textbf{23.5 ± 0.31} & \textbf{10.0 ± 0.69} \\ \bottomrule
\end{tabular}%
\caption{Evaluation results on the CHAIR benchmark using the MSCOCO dataset (val2014 split). We conduct experiments with three different sets of 500 images, each selected by random seeds. The reported value is the mean of the results from the three different seeds, with the $\pm$ symbol representing the standard deviation.}
\label{tab:CHAIR}
\end{table*}

Note that, Equation~\ref{eq:3_contrastive_decoding} is similar to the contrastive decoding methods used in VCD~\cite{leng2024mitigating} and HALC~\cite{chen2024halc}. 
However, our method is distinguished from existing approaches by directly capturing visual contrastive signals from the hallucinations visualized by the T2I generative model.

\section{Experiments}
\label{sec:4_experiments}

\noindent
\textbf{Benchmarks.}
To evaluate the performance of our method, we conduct experiments on three benchmarks to evaluate the mitigation of hallucinations and two general-purpose benchmarks to assess the general performance of the MLLM: 
\begin{itemize}
    \item Hallucination: \textit{CHAIR}~\cite{rohrbach2018object}, \textit{HallusionBench}~\cite{guan2024hallusionbench}, and \textit{Polling-based Object Probing Evaluation (POPE)}~\cite{li2023evaluating}
    \item General-purpose: \textit{MLLM Evaluation (MME)}~\cite{fu2023mme} and  \textit{LLaVA-Bench}~\cite{liu2024visual}
\end{itemize}
Detailed information on these benchmarks can be found in the Appendix.

\noindent
\textbf{Backbones.}
To evaluate our method, we utilize three well-known MLLMs with publicly available checkpoint weights: LLaVA-1.5~\cite{liu2024improved}, mPLUG-Owl2~\cite{ye2024mplug}, and MiniGPT-4~\cite{zhu2024minigpt}.

\noindent
\textbf{Compared Methods.}
Our method is designed to replace existing decoding methods used in the LLM component, and therefore, we compare it against baselines such as Greedy Search, Nucleus Sampling~\cite{holtzman2020curious}, and Beam Search (beam=5).
We also evaluate our method's effectiveness against other decoding methods in hallucination mitigation, including OPERA~\cite{huang2024opera}, VCD~\cite{leng2024mitigating}, and HALC~\cite{chen2024halc}. 
We use the same hyperparameters borrowed from the original papers of the compared methods to ensure a fair comparison.

\noindent
\textbf{Implementation Details.}
We utilize the Hyper-SDXL~\cite{ren2024hyper} T2I model for image generation. 
Specifically, in all experiments, unless otherwise noted, we use the Step 1 generation results of Hyper-SDXL model. 
The maximum length of text queries that the T2I model could accept is 77 tokens, which is too short to process the captions generated by MLLM. 
To address this, we leverage Compel~\cite{compel}, which allows for processing more than 77 tokens. 
We set the maximum token count for the caption generation to 256 and use Nucleus sampling with a temperature of 0.7 and a top-$p$ of 0.9 to generate the images. 
The query used in this process is ``Please describe this image in detail.'' 
We set the number of generated images, $n$, to 4, producing four images based on distinct captions generated using different random seeds.
For contrastive decoding, we follow~\cite{li2023contrastive} using adaptive plausibility constraint to contrast only meaningful tokens.
The plausibility constraint hyperparameter \(\lambda\) is set to 0.1.
We also set \(\alpha\), which controls the degree of contrastive emphasis, to 1 for captioning-based metrics such as CHAIR and LLaVA-Bench, and to 0.1 for VQA metrics, including POPE, HallusionBench, and MME. 
To generate responses, we use a greedy decoding approach for all methods.
For CHAIR, we sample three different sets of images using different random seeds and assess the performance using the mean and standard deviation of these results.

\subsection{Experimental Results}\label{4_2_experimental_results}

\textbf{Results on CHAIR.}
We report our evaluation results on the CHAIR~\cite{rohrbach2018object} benchmark in Table~\ref{tab:CHAIR}. 
Our assessment includes basic decoding strategies—Greedy search, Nucleus sampling, and Beam search—along with three state-of-the-art approaches—VCD~\cite{leng2024mitigating}, OPERA~\cite{huang2024opera}, and HALC~\cite{chen2024halc}.
Our method achieves the best performance on the CHAIR$_S$ metric across all three backbone models (LLaVA-1.5, mPLUG-Owl2, and MiniGPT-4). 
Remarkably, it significantly improves the CHAIR$_S$ score compared to both the basic decoding strategies and the state-of-the-art methods, highlighting its superior ability to mitigate hallucinations.
In terms of the CHAIR$_I$ metric, our method consistently ranks either first or second across all backbone models.
These results demonstrate that our method both excels in reducing the total number of hallucinations throughout entire sentences and minimizes the number of hallucinated objects across all evaluated image sets.

\begin{table}[t]
\centering
\small
\begin{tabular}{@{}c|c|c@{}}
\toprule
Method           & Figure Acc (fAcc) & All Acc (aAcc) \\ \midrule
Greedy Search           & \underline {22.2}        & 50.1           \\
Nucleus Sampling & 17.8              & 46.2           \\
Beam Search       & 19.1              & 48.4           \\ \midrule
VCD              & 21.7              & 47.5           \\
OPERA            & 20.9              & 49.9           \\
HALC             & 21.7              & \underline {50.6}     \\ \midrule
Ours             & \textbf{23.5}     & \textbf{50.8}  \\ \bottomrule
\end{tabular}
\caption{Evaluation results on HallusionBench. We report Figure Acc and All Acc using LLaVA-1.5.}
\label{tab:HallusionBench}
\end{table}
\begin{table}[t]
    \centering
    \small
    \setlength{\tabcolsep}{3.75pt}
    \begin{tabular}{c|c|c|c|c}
        \toprule
        {Method} & LLaVA-1.5 & mPLUG-Owl2 & MiniGPT-4 & Average \\ 
        \midrule
        VCD & \underline{82.8} & 81.6 & 59.8 & 74.7 \\
        OPERA & \textbf{83.0} & \underline{83.3} & 66.1 & \underline{77.4} \\
        HALC & 50.6 & \textbf{83.4} & \underline{69.7} & 67.9 \\ \midrule
        Ours & \textbf{83.0} & 83.0 & \textbf{69.9} & \textbf{78.6} \\
        \bottomrule
    \end{tabular}
    \caption{Evaluation results on the POPE benchmark using the MSCOCO dataset (val2014 split).}
    \label{tab:POPE}
\end{table}

\begin{table}[t]
\centering
\small
\begin{tabular}{@{}c|ccc@{}}
\toprule
\multirow{2}{*}{Method} & \multicolumn{2}{c|}{Category}                          & \multirow{2}{*}{Total}      \\ \cmidrule(lr){2-3} 
                        & \multicolumn{1}{c|}{Perception}      &  \multicolumn{1}{c|}{Cognition}       &                        \\ \midrule
Greedy Search                 & \multicolumn{1}{c|}{1472.5}          & \multicolumn{1}{c|}{303.9}           & \underline{1776.4}   \\
Nucleus Sampling        & \multicolumn{1}{c|}{1203.4}          & \multicolumn{1}{c|}{311.1}           & 1514.5   \\
Beam Search              & \multicolumn{1}{c|}{\underline{1478.0}}    & \multicolumn{1}{c|}{287.5}           & 1765.5   \\ \midrule
VCD                     & \multicolumn{1}{c|}{1326.7}          & \multicolumn{1}{c|}{\textbf{374.6}}  & 1701.3   \\
OPERA                   & \multicolumn{1}{c|}{1456.9}          & \multicolumn{1}{c|}{306.4}           & 1763.3   \\
HALC                    & \multicolumn{1}{c|}{887.7}           & \multicolumn{1}{c|}{269.6}           & 1157.3   \\ \midrule
Ours                    & \multicolumn{1}{c|}{\textbf{1487.6}} & \multicolumn{1}{c|}{306.1}           & \textbf{1793.7}   \\ \bottomrule
\end{tabular}
\caption{Evaluation results on the MME using LLaVA-1.5.}
\label{tab:MME-LLaVA_total}
\end{table}

\begin{table}[t]
\centering
\setlength{\tabcolsep}{4pt}
\begin{tabular}{@{}c|cccc@{}}
\toprule
Method           & Complex       & Conv          & Detail        & All           \\ \midrule
Greedy Search           & 82.0          & 47.3          & 64.1          & 67.0          \\
Nucleus Sampling & 76.2          & 41.2          & 52.6          & 59.9          \\
Beam Search       & 83.9          & 58.7          & 58.8          & 70.0          \\ \midrule
VCD              & 79.9          & 53.5          & 56.3          & 66.2          \\
OPERA            & 78.7          & 53.0          & 58.3          & 66.0          \\
HALC             & 55.8          & 31.1          & 50.4          & 47.1          \\ \midrule
Ours             & \textbf{84.2} & \textbf{63.5} & \textbf{64.8} & \textbf{73.3} \\ \bottomrule
\end{tabular}
\caption{Evaluation results on LLaVA-Bench using LLaVA-1.5.}
\label{tab:llava-bench}
\end{table}

\begin{table*}[t]
\centering
\begin{tabular}{@{}c|c|cc|cc|cc@{}}
\toprule
\multirow{2}{*}{T2I Model} & \multirow{2}{*}{CLIPScore ↑} & \multicolumn{2}{c|}{LLaVA-1.5} & \multicolumn{2}{c|}{mPLUG-Owl2} & \multicolumn{2}{c}{MiniGPT-4} \\ \cmidrule(l){3-8} 
            &       & CHAIR$_S$ ↓ & CHAIR$_I$ ↓ & CHAIR$_S$ ↓ & CHAIR$_I$ ↓ & CHAIR$_S$ ↓ & CHAIR$_I$ ↓ \\ \midrule
Hyper-SD1.5 & 30.87 & 20.2       & 6.6        & 19.4       & 6.4        & 28.2       & 11.8       \\
SDXL-Turbo  & 32.33 & 18.8       & 6.6        & 20.2       & 6.68       & 25.2       & 9.9        \\
Hyper-SDXL  & 32.85 & 17         & 5.6        & 17         & 5.3        & 24.4       & 10.0       \\ \bottomrule
\end{tabular}
\caption{Our performance when differentiating the T2I models for visualizing hallucinations. We generate captions with nucleus sampling and set max new token for 64 and generate the image with those captions. Inference step for diffusion set to be all 1.}
\label{tab:VariousGenModel}
\end{table*}
\begin{table}[t]
\centering
\small
\begin{tabular}{@{}c|c|c|c@{}}
\toprule
Captioning by                & LLaVA-1.5       & mPLUG-Owl2      & MiniGPT-4        \\ \midrule
\multicolumn{4}{c}{CHAIR$_S$ ↓}                                              \\ \midrule
Greedy Search                & 19.4            & 19.4            & 27.2            \\
Nucleus Sampling      & 18.8            & 15.2            & 24.4            \\ \midrule
\multicolumn{4}{c}{CHAIR$_I$ ↓}                                              \\ \midrule
Greedy Search               & 6.6             & 6.4             & 11.6            \\
Nucleus Sampling      & 6.7             & 5.1             & 10.3            \\ \bottomrule
\end{tabular}
\caption{Comparison of performance using a single image (\(n=1\)) generated by two different decoding strategies, Greedy search and Nucleus sampling.}
\label{tab:Decoding_strategy for_img_generation}
\end{table}

\noindent
\textbf{Results on HallusionBench.}
In Table~\ref{tab:HallusionBench}, we present the evaluation results for the visual dependent category of the HallusionBench~\cite{guan2024hallusionbench} benchmark.
HallusionBench is evaluated with the assistance of GPT-4V, which incurs significant costs; therefore, we conduct experiments using only the LLaVA-1.5~\cite{liu2024improved} backbone.
Our method demonstrates superior performance in Figure Accuracy (fAcc), outperforming all baseline decoding strategies (Greedy Search, Nucleus Sampling, Beam Search) as well as state-of-the-art techniques (VCD, OPERA, HALC). 
This indicates that our model effectively interprets the visual details of images when responding to visually dependent questions, indicating its ability to mitigate hallucinations by providing responses that closely align with the given visual content.
Furthermore, our method achieves the highest performance on the All Accuracy (aAcc) metric, which measures overall accuracy across all questions within the visual dependent category, demonstrating its effectiveness in handling a wide range of visually dependent queries.

\noindent
\textbf{Results on POPE.}
Table~\ref{tab:POPE} reports the evaluation results on the POPE~\cite{li2023evaluating} benchmark using the MSCOCO~\cite{lin2014microsoft} dataset (val2014 split). 
We present the average F1-scores across the three POPE question splits—Random, Popular, and Adversarial—for three different backbone models. Detailed performances on each POPE question split are in the Appendix.

Our method achieve a new SOTA performance on MiniGPT-4, and demonstrate performance comparable to existing techniques on LLaVA-1.5 and mPLUG-Owl2. 
In terms of average performance across all backbones, our method outperforms previous techniques. 
This indicates that our approach consistently delivers strong performance across various backbones.

While we achieves overall strong performance on this benchmark, the performance improvements across different backbone models are relatively modest. 
This might be because the POPE question split does not fully align with the types of hallucinations that T2I models generate. 
POPE questions, which ask, ``Is this [object] in this image?'' sample objects randomly, popularly, or adversarially. 
Meanwhile, our method visualizes hallucinations in captions generated by prompts like ``Please describe this image in detail.'' 
As a result, T2I model may visualize the objects unrelated to the actual POPE questions which limits our method's effectiveness. 
This limitation will be explored further through a qualitative analysis of POPE samples later in this section.

\noindent
\textbf{Results on MME.}
In Table~\ref{tab:MME-LLaVA_total}, we present the evaluation results on the MME benchmark using the LLaVA-1.5 backbone. 
Due to space limitations, we focus on the performance in the two main categories of the MME benchmark: Perception and Cognition. 
Scores for the subcategories are provided in the Appendix.
Our method outperforms all others in the Perception category, demonstrating its effectiveness in accurately interpreting and processing visual information across various tasks. 
This strong performance indicates that our model is particularly well-suited for visual tasks, making it highly effective for applications that require precise visual understanding.
In the Cognition category, our method demonstrates competitive performance, comparable to OPERA and superior to HALC, further underscoring the versatility and robustness of our approach. 
While VCD excels in cognitive tasks, our method achieves stronger overall performance when both the Perception and Cognition categories are considered together. 
This suggests that our model provides a more comprehensive and effective solution across diverse tasks. 
Its balanced and reliable performance in both visual and cognitive challenges makes it an adaptable solution for a wide range of applications.

\noindent
\textbf{Results on LLaVA-Bench.}
Table~\ref{tab:llava-bench} shows the experimental results on the LLaVA-Bench, which verify whether the language model capabilities are preserved. 
For this evaluation, we uses the LLaVA-1.5 backbone. 
Our method outperforms existing techniques across all categories: complex reasoning, conversation, and detailed description.
These results demonstrate that our method effectively mitigates hallucinations while also enhancing the performance of the MLLM.

\begin{figure}[t]
    \centering
        \begin{subfigure}{0.49\columnwidth}
            \includegraphics[width=\textwidth]{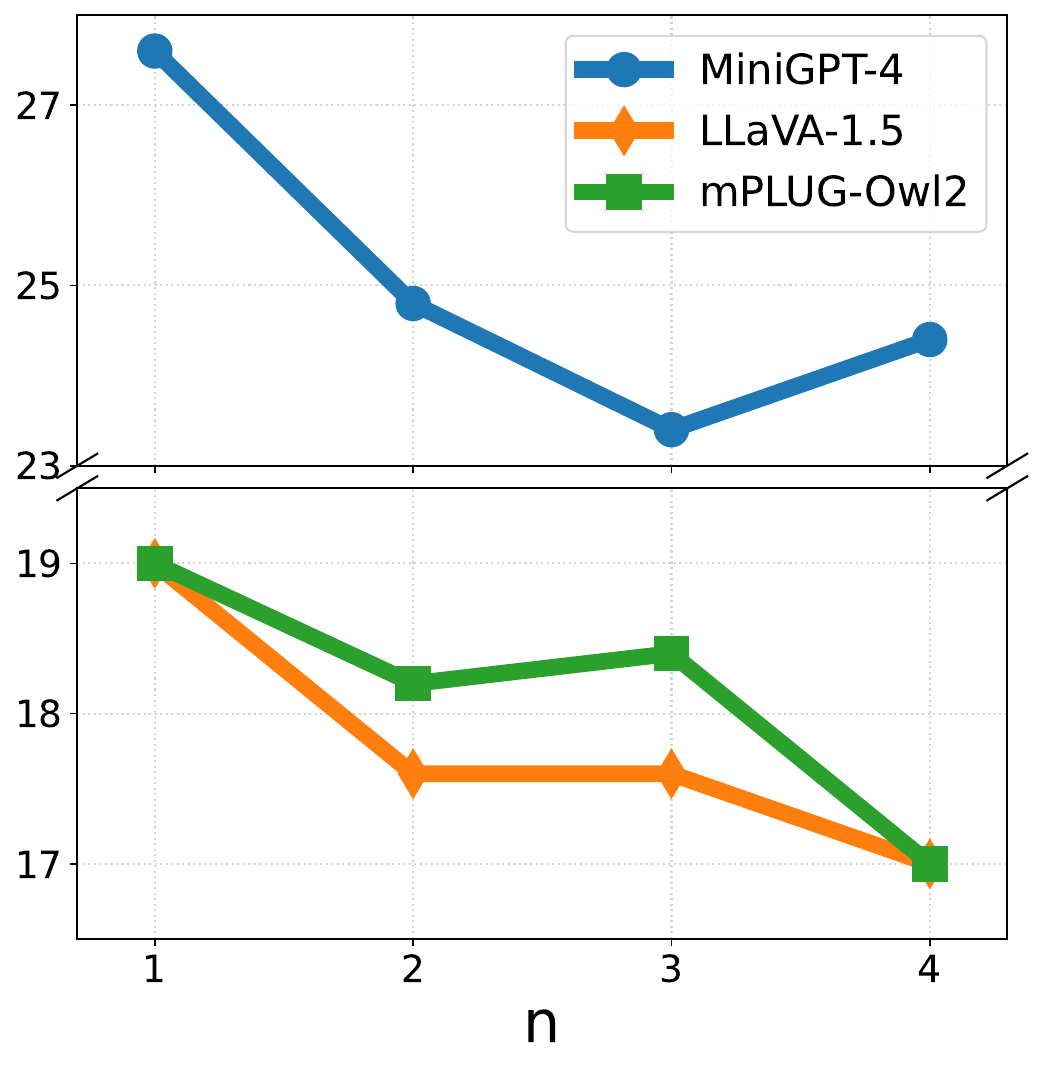}
            \caption{CHAIR$_S$ Score}
           \label{fig:4a_analysis_chairs}
         \end{subfigure}
        \begin{subfigure}{0.49\columnwidth}
         \includegraphics[width=\textwidth]{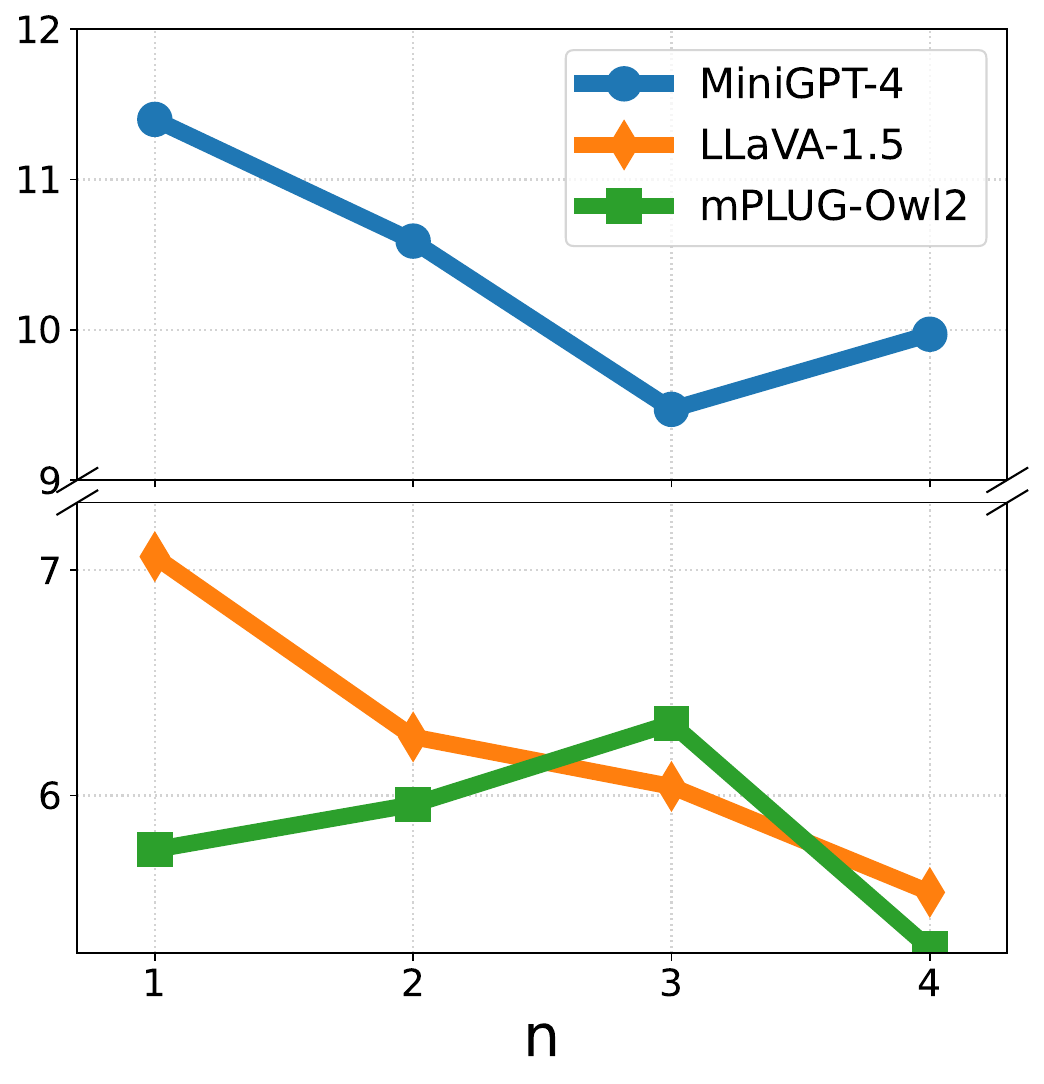}
         \caption{CHAIR$_I$ Score}
         \label{fig:4b_analysis_chairi}
        \end{subfigure}
    \caption{Effect of the number of images with different captions.}
    \label{fig:4_analysis}
\end{figure}

\subsection{Analysis and Discussion}\label{subsec:4_3_analysis_and_discussion}

\textbf{Diversity of Generated Captions and Images.}
Although T2I models have made significant advancements, they still struggle to generate images that perfectly align with the given captions~\cite{Ruiz2023CVPR}.
To address these limitations, we increase the coverage of hallucination visualization by generating diverse images. 
Specifically, we use Nucleus sampling, which is known for producing more varied responses than Greedy search, to generate multiple captions. 
These captions are then utilized to generate images.

To evaluate the effectiveness of this strategy, we analyze how caption diversity impacts hallucination reduction. 
First, we compare the CHAIR scores of the final responses when using Greedy search and Nucleus sampling during the image generation stage.
In this experiment, we limit the number of generated images to one and compare which decoding strategy performs better. 
As shown in Table~\ref{tab:Decoding_strategy for_img_generation}, Nucleus sampling outperforms Greedy search, demonstrating its potential to generate more diverse captions.
Furthermore, in Figure~\ref{fig:4_analysis}, we investigate how the number of generated images from different captions using Nucleus sampling affects CHAIR scores.
We observe that the number of images $n$ increases, both CHAIR$_S$ and CHAIR$_I$ scores improve, confirming that using multiple reconstructed images, rather than a single image, is more effective for improving performance. 
These findings validate our design choice of utilizing Nucleus sampling and multiple captions for image generation.

\noindent
\textbf{Impacts of Image Generation Quality.}
To investigate the impact of generated image quality on hallucination mitigation, we evaluate the performance of our method using various text-to-image (T2I) models. 
Table~\ref{tab:VariousGenModel} presents the generation quality (CLIPScore) of the T2I models alongside their corresponding CHAIR scores. 
We compare three T2I models: Hyper-SD1.5~\cite{ren2024hyper}, SDXL-Turbo~\cite{sauer2023adversarial}, and Hyper-SDXL~\cite{ren2024hyper}, with the inference step fixed at 1.

The results indicate a clear trend: as the CLIPScore improves, so does the CHAIR score.
Notably, SDXL-Turbo consistently outperforms Hyper-SD1.5 across all backbones, except for mPLUG-Owl2. 
Moreover, Hyper-SDXL significantly outperforms Hyper-SD1.5 in all cases. 
These findings suggest that using higher-quality T2I models, which are better aligned with the original captions, can more effectively mitigate hallucination issues. 
Consequently, we believe that as more advanced T2I models are developed, the performance of our method will continue to improve.

\begin{figure}[t]  
    \centering
    \includegraphics[width=0.4\textwidth]{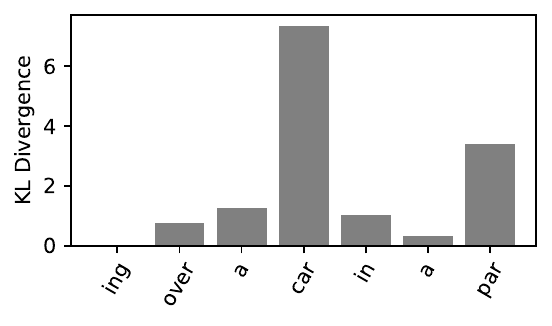}
    \caption{KL divergence between output distributions across each decoding step when the MLLM is provided with the images and caption from Figure~\ref{fig:6_qualitative_samples} (a). The KL divergence is significantly elevated for the hallucinated token ``car''.}  
    \label{fig:5_qualitative_kl} 
\end{figure}

\begin{figure*}[t]
    \centering
        \begin{subfigure}{0.45\textwidth}
            \includegraphics[width=\textwidth]{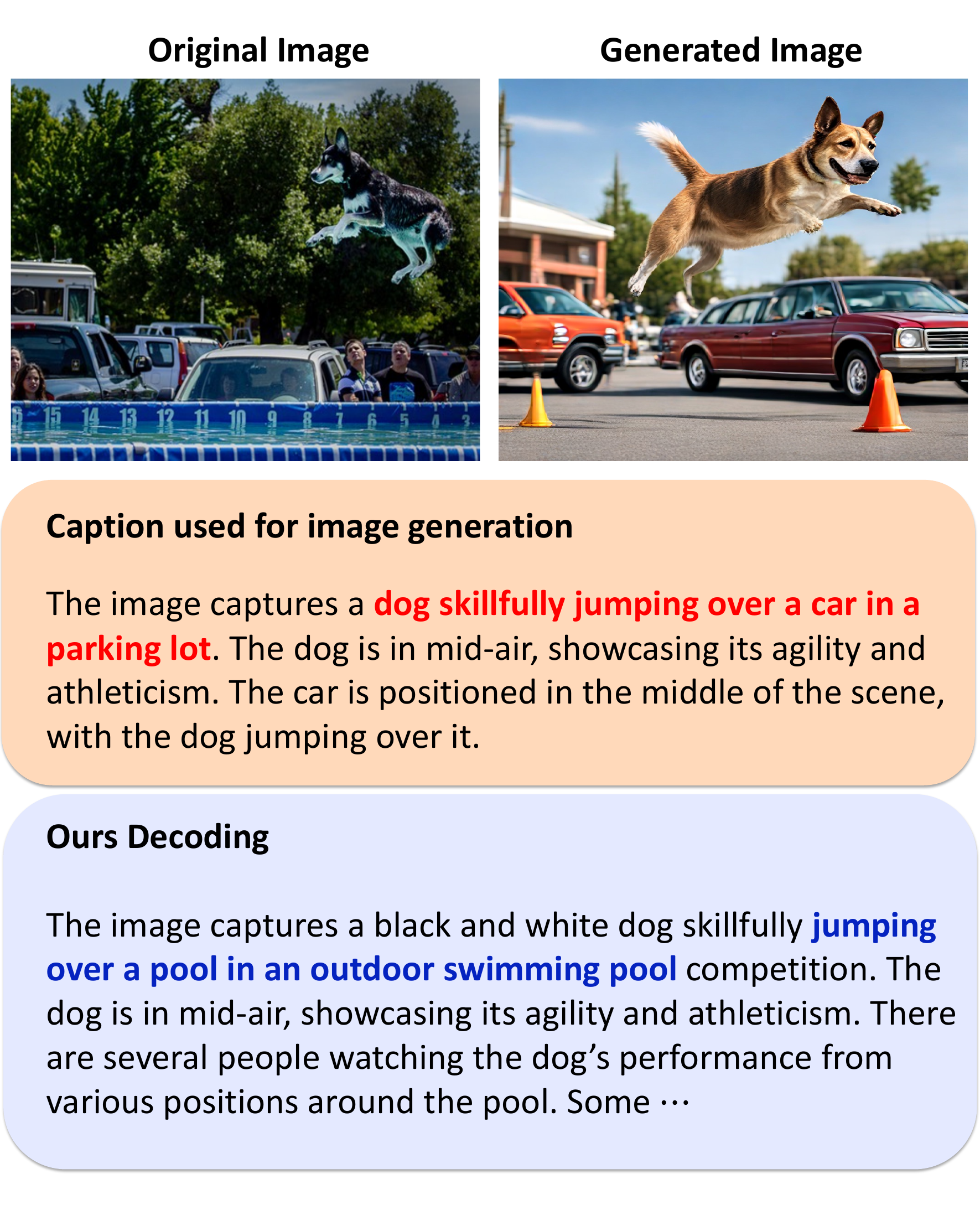}
            \caption{}
           \label{fig:4a_qualitative}
         \end{subfigure}\hspace{0.05\textwidth}
        \begin{subfigure}{0.45\textwidth}
         \includegraphics[width=\textwidth]{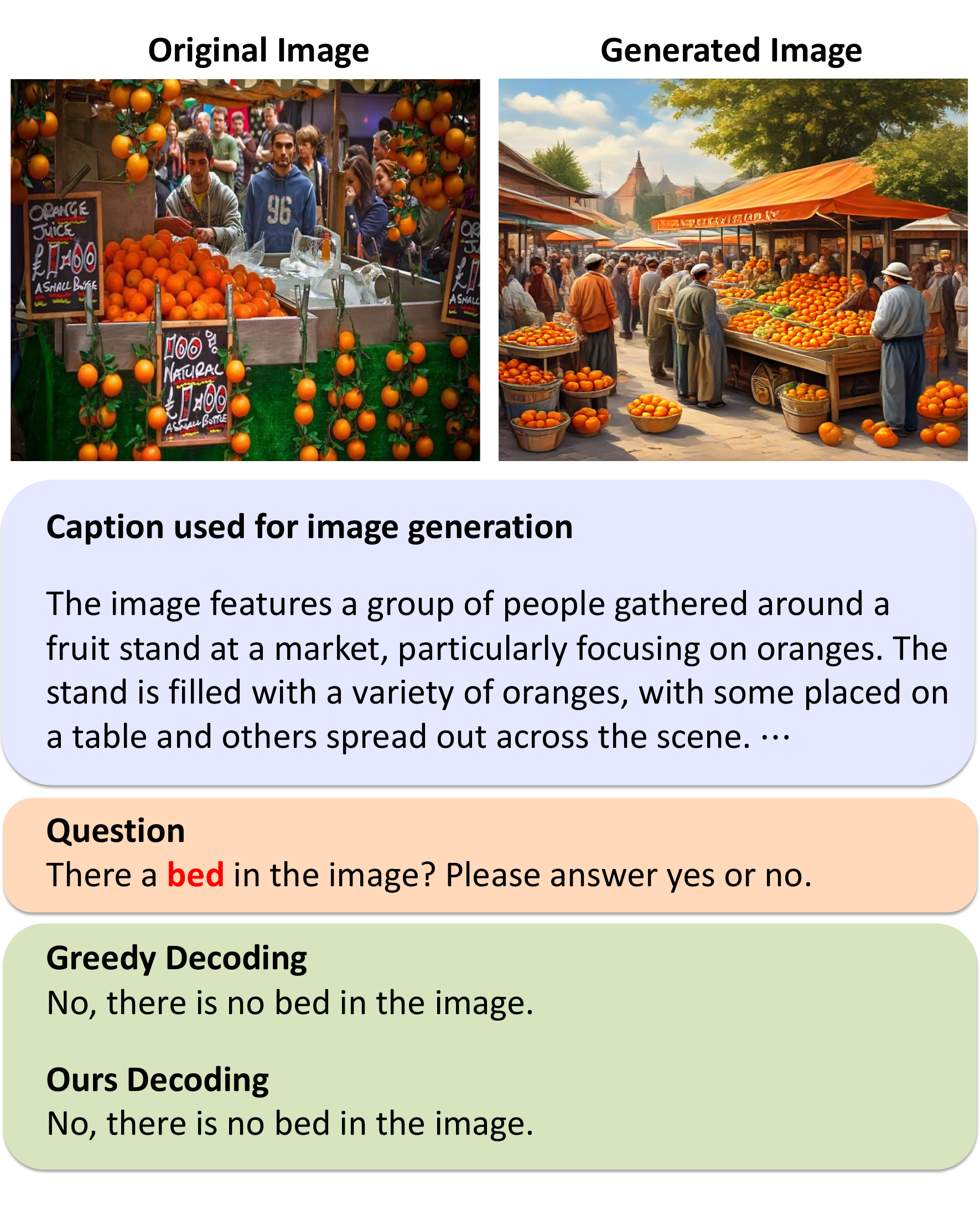}
         \caption{}
         \label{fig:4b_limitation}
        \end{subfigure}
    \caption{Qualitative samples using LLaVA-1.5 for backbone model. (a) shows an example that T2I model faithfully generate the images that depicts the hallucination in the caption. (e.g., jumping over a car) (b) is an example of our limitation in VQA tasks, which there can be a misalignment between visualized hallucination and actual main subject of question.}
    \label{fig:6_qualitative_samples}
\end{figure*}

\noindent
\textbf{Qualitative Analysis.}
Figure~\ref{fig:5_qualitative_kl} shows the KL divergence between output distributions at each decoding step when the images and caption from Figure~\ref{fig:6_qualitative_samples} (a) are provided to the MLLM. 
We observe that the KL divergence is high for the hallucinated token \textit{car}, while non-hallucinated tokens exhibit lower KL divergence. 
This indicates that the generated image can produce visual contrastive signals for hallucinated tokens when compared to the original image. 
This supports our argument that the differences between the original and generated images are primarily influenced by the hallucinated tokens.

To more clearly demonstrate how our method mitigates multimodal hallucinations, we present an example in Figure~\ref{fig:6_qualitative_samples} (a), illustrating the process from the initial hallucinated caption to the generated image, followed by the contrastive decoding result. 
Specifically, for an image of a dog jumping into a pool, the MLLM incorrectly describes the scene as ``a dog jumping over a car in a parking lot.'' 
Using this caption, the T2I model generates a reconstructed image that faithfully visualized the hallucinated content. 
By contrasting the distributions of the reconstructed and original images during decoding, our method effectively reduces hallucinations.

\noindent
\textbf{Limitations.}
One of the key limitations of our approach is its strong dependence on T2I generation models.
This reliance may hinder effectiveness in tasks like VQA, where the generated captions can sometimes contain hallucinations that deviate significantly from the specific question.
This limitation is particularly evident in our experiments with the POPE benchmark, where the performance gain is not as significant as expected.
Regarding questions about the presence of specific objects, if the object in question is not related to the hallucinations generated by the caption, visualizing with a T2I model may not sufficiently reflect the information needed for the VQA task.
In Figure~\ref{fig:6_qualitative_samples} (b), a question about the presence of a bed in an original image where people are looking at fruits might not be well served by the reconstructed image.
This indicates the effectiveness of our method may decrease for certain type of questions.

Currently, our technique employs a fixed prompt for image captioning. 
However, we believe that adapting the prompt to respond more specifically to the given question could mitigate this issue. 
We plan to explore this adaptive approach in future work.

\section{Conclusion}\label{sec:5_conclusion}

In this paper, we presented ConVis, a novel contrastive decoding method designed to mitigate hallucinations in MLLMs.
By utilizing a T2I generation model, our approach effectively visualizes hallucinations and contrasts probability distributions between the original and reconstructed images. 
This process allows for the penalization of hallucinated content during the decoding phase, all without the need for additional data or model retraining.

Our extensive experiments across five benchmarks, including CHAIR, HallusionBench, and LLaVA-Bench, demonstrated that ConVis consistently reduces hallucinations while preserving the core language model capabilities of MLLMs.
The method achieves competitive or superior performance compared to existing techniques in various categories, validating its effectiveness in enhancing the reliability of MLLM outputs.
\bibliography{aaai25}

\clearpage
\appendix
\section{Appendix}

\subsection{Benchmarks}
\setcounter{table}{0}
\renewcommand{\thetable}{A\arabic{table}}

\noindent
In this appendix, we provide additional details into the benchmarks referenced in the main paper.
To evaluate hallucinations, we employ the following five benchmarks:

\textbf{CHAIR}~\cite{rohrbach2018object} evaluates how well the generated captions align with the content of the given image. 
CHAIR consists of two versions: CHAIR\_S, which measures the inaccuracies at the sentence level, and CHAIR\_I, which evaluates at the object level within the sentence by comparing the number of false objects to the total number of objects. 
For evaluation, we use the val2014 split of the MSCOCO~\cite{lin2014microsoft} dataset, which includes annotations for 80 object categories. 
We randomly select 500 images from the entire dataset and used the prompt ``Please describe this image in detail.'' for the MLLM.

\textbf{HallusionBench}~\cite{guan2024hallusionbench} is a hallucination evaluation benchmark designed to assess whether a model ignores visual context and relies solely on language priors (Language Hallucination) or exhibits the opposite phenomenon (Visual Illusion). 
The questions in HallusionBench are divided into two main categories, one of which is the Visual Dependent (VD) category. 
In this category, pairs of similar but different images are presented, and the same question is asked for each pair. 
The questions are presented in a VQA format with binary ground truth (GT) answers. 
Accuracy is calculated using GPT-4V by determining whether the model’s responses are similar to, different from, or difficult to compare with the answers generated by GPT-4V. 
Since this paper focuses on preventing MLLMs from generating hallucinated information based on a given image, we specifically conduct experiments on the Visual Dependent category.

\textbf{Polling based Object Probing Evaluation (POPE)}~\cite{li2023evaluating} is a VQA-based metric proposed to assess hallucinations in MLLMs. 
This metric evaluates the MLLM’s response to the prompt ``Is [object] is in this image?'' To emphasize that this is a binary VQA task, we appended the prompt with ``Please answer yes or no.''
To select objects referenced in the question prompt, we followed three different sampling options: random, popular, and adversarial. 
We evaluated performance across all sampling options.

\textbf{MLLM Evaluation (MME)}~\cite{fu2023mme} evaluates the capabilities of MLLMs, dividing the evaluation into two major categories: perception and cognition. 
The perception category includes fine-grained tasks such as existence, count, location, rough color, poster, celebrity, scene, landmark, artwork identification, and OCR. 
The cognition category includes tasks like commonsense reasoning, numerical calculations, text translation, and code reasoning. 
All questions in this benchmark are structured to be answered with a simple yes or no.

Using the \textbf{LLaVA-Bench}~\cite{liu2024visual}, we further demonstrated how well our proposed method maintains the language model performance.
This benchmark involves posing various situational questions, such as dialogue, detailed descriptions, and complex reasoning, to randomly selected images from the MSCOCO val2014 dataset. 
A total of 60 questions are used to assess whether the model faithfully follows the instructions. 
The generated answers are evaluated by comparing them to the responses of a text-only GPT-4 model.

\subsection{Additional Implementation Details and Experimental Results}
We present further implementation details and experimental results that were omitted from the main paper due to space limitations. Table~\ref{tab:max_new_token} outlines the maximum lengths set for response generation. Additionally, Table~\ref{tab:POPE_All} provides the complete evaluation results on the POPE benchmark using the MSCOCO dataset, including analyses of Random, Popular, and Adversarial scenarios across three MLLM backbones. Finally, Table~\ref{tab:MME_LLaVA} offers a full comparison of category performance for the MME benchmark in LLaVA-1.5.

\begin{table}[t]
\centering
\begin{tabular}{@{}c|c@{}}
\toprule
Benchmark      & Max New Tokens \\ \midrule
CHAIR          & 64             \\
HallusionBench & 64             \\
POPE           & 16             \\
MME            & 128            \\
LLaVA-Bench    & 512            \\ \bottomrule
\end{tabular}%
\caption{Maximum number of generated tokens utilized in the response generation for each benchmark experiment.}
\label{tab:max_new_token}
\end{table}

\begin{table*}[t]
    \centering
    \small\resizebox{\textwidth}{!}{%
    \begin{tabular}{c|ccc|ccc|ccc|c}
        \toprule
    \multirow{2}{*}{Method} & \multicolumn{3}{c|}{LLaVA-1.5} & \multicolumn{3}{c|}{mPLUG-Owl2} & \multicolumn{3}{c|}{MiniGPT-4} & \multirow{2}{*}{Average} \\ 
    \cmidrule(lr){2-4} \cmidrule(lr){5-7} \cmidrule(lr){8-10}
    & Random & Popular & Adversarial & Random & Popular & Adversarial & Random & Popular & Adversarial &  \\
    \midrule
        Greedy & 84.6 & \underline{83.4} & \underline{81.3} & \underline{85.8} & \underline{83.5} & 80.3 & \underline{74.1} & \underline{68.2} & \textbf{67.1} & \textbf{78.7} \\
        Sample (Nucleus) & 78.7 & 77.0 & 76.2 & 82.5 & 79.5 & 76.9 & 60.5 & 61.6 & 57.3 & 72.24 \\
        Beam (n=5) & \underline{85.0} & \textbf{83.7} & \textbf{81.5} & 85.5 & \underline{83.5} & \textbf{80.7} & 71.0 & 67.6 & 64.6 & 78.12 \\ \midrule
        VCD & \textbf{85.3} & 82.9 & 80.1 & 84.7 & 81.8 & 78.4 & 61.5 & 59.3 & 58.7 & 74.74 \\
        OPERA & 84.4 & \underline{83.4} & 81.2 & \underline{85.8} & 83.5 & \underline{80.5} & 69.3 & 65.7 & 63.2 & 77.44 \\
        HALC & 50.8 & 50.6 & 50.4 & \textbf{86.0} & \textbf{83.6} & \underline{80.5} & \textbf{74.3} & 68.1 & \underline{66.8} & 67.90 \\ \midrule
        Ours & 84.7 & 83.2 & 81.1 & 85.6 & 83.1 & 80.2 & \textbf{74.3} & \textbf{68.3} & \textbf{67.1} & \underline{78.62} \\
    \bottomrule
    \end{tabular}%
    }
    \caption{Full report of evaluation on the POPE benchmark using the MSCOCO dataset (val2014 split).}
    \label{tab:POPE_All}
\end{table*}

\begin{table*}[t]
\centering
\resizebox{\textwidth}{!}{%
\begin{tabular}{@{}c|cccccccccc|cccc@{}}
\toprule
Method &
  Existence &
  Count &
  Position &
  Color &
  Posters &
  Celebrity &
  Scene &
  Landmark &
  Artwork &
  OCR &
  Common &
  Numerical &
  Text Trans &
  Code Reas \\ \midrule
Greedy &
  \textbf{195.0} &
  133.3 &
  \underline{133.3} &
  \textbf{155.0} &
  \underline{138.0} &
  128.5 &
  \underline{153.5} &
  153.2 &
  \textbf{125.0} &
  \textbf{132.5} &
  \underline{121.4} &
  37.5 &
  82.5 &
  62.5 \\
Sample (Nucleus) &
  \underline{180.0} &
  136.6 &
  116.6 &
  138.3 &
  122.7 &
  \underline{131.4} &
  146.2 &
  145.2 &
  109.2 &
  100.0 &
  \textbf{122.1} &
  \textbf{85.0} &
  \textbf{92.5} &
  \textbf{75.0} \\
Beam (n=5) &
  \textbf{195.0} &
  153.3 &
  \underline{133.3} &
  \textbf{155.0} &
  135.7 &
  126.7 &
  152.7 &
  152.5 &
  122.5 &
  \underline{130.0} &
  116.4 &
  40.0 &
  \underline{87.5} &
  62.5 \\ \midrule
VCD &
  \textbf{195.0} &
  \textbf{163.3} &
  \textbf{138.3} &
  \textbf{155.0} &
  \underline{138.0} &
  128.5 &
  152.7 &
  \underline{154.0} &
  \underline{123.0} &
  \underline{130.0} &
  115.0 &
  45.0 &
  65.0 &
  62.5 \\
OPERA &
  165.0 &
  101.6 &
  98.3 &
  \underline{153.3} &
  116.6 &
  107.9 &
  135.2 &
  125.7 &
  109.5 &
  90.0 &
  108.5 &
  \underline{67.5} &
  67.5 &
  \underline{67.5} \\
HALC &
  110.0 &
  78.3 &
  90.0 &
  100.0 &
  60.2 &
  69.4 &
  109.5 &
  98.5 &
  101.7 &
  70.0 &
  92.1 &
  50.0 &
  77.5 &
  50.0 \\ \midrule
Ours &
  \textbf{195.0} &
  \underline{158.3} &
  \underline{133.3} &
  \textbf{155.0} &
  \textbf{143.2} &
  \textbf{139.7} &
  \textbf{153.8} &
  \textbf{155.3} &
  121.5 &
  \textbf{132.5} &
  118.6 &
  45.0 &
  \underline{87.5} &
  55 \\ \bottomrule
\end{tabular}%
}
\caption{Evaluation results on the MME benchmark using LLaVA-1.5 for MLLM backbone, conducted across 10 subcategories focused on perception and 4 subcategories focused on cognition.}
\label{tab:MME_LLaVA}
\end{table*}

\end{document}